\documentclass[
]{ceurart}

\usepackage{float}
\usepackage{latexsym}
\usepackage{graphicx}
\usepackage{subcaption}
\usepackage{booktabs}
\usepackage{microtype}
\usepackage{amsmath}
\usepackage{enumitem}
\usepackage{todonotes}
\usepackage{caption}
\begin{document}

\copyrightyear{2026}
\copyrightclause{Copyright for this paper by its authors.
  Use permitted under Creative Commons License Attribution 4.0
  International (CC BY 4.0).}

\conference{CLiC-it 2026: Twelfth Italian Conference on Computational Linguistics,
  September 14--16, 2026, Palermo, Italy}

\title{Language, Language Models, and What We're Talking About}



\author{Malvina Nissim}[%
orcid=0000-0001-5289-0971,
email=m.nissim@rug.nl,
url=https://malvinanissim.github.io/,
]
\address{Center for Language and Cognition, Faculty of Arts, University of Groningen\\Oude Kijk in Het Jatstraat 26, 9712~EJ Groningen, The Netherlands}



\begin{abstract}
Language models are commonly discussed as technical artefacts, but they are obviously shaped by the linguistic worlds conveyed by data during their training. Using Italian language models as evidence, I want to bring attention to the nature of the systems which result from training and specialising models on translated and synthetic data, and further curating them, and to the meaning of testing them on equally unnatural data. Are these eventually models of Italian? Are they models of language? Does NLP still care about \textit{language}? These questions yield another, more concrete question: what language do we actually want language models to produce? I argue that this question cannot be answered if we do not first consider a clearer distinction between language models designed as technical products and language models designed as tools for studying language itself. The answers then might be diverse, the languages we are talking about might be diverse, and the picture might not be as pessimistic as we fear.
\end{abstract}


\footnotetext[1]{{While the process is obviously well known to the NLP community and might not need any visualisation, I care to include it as the different stages are core to the paper's discussion. Also, I am very much aware of the simplification realised in this picture, considering extensions and modifications in current systems as mentioned in the text, it still serves as a good reference for the structure and discussion of this paper. As my drawing skills are very limited and rudimentary, this diagram, and all figures in this paper, are generated using either Claude or Nano Banana unless otherwise specified.}\smallskip}

\maketitle

\section{The State of Things}

When we talk about a "\textit{Large Language Model}" (LLM), we usually refer to a technical artefact: a neural network which nowadays, more often than not, is based on the Transformer architecture \citep{vaswani2017attention} 
\begin{wrapfigure}{r}{0.5\textwidth}
\includegraphics[width=0.45\textwidth]{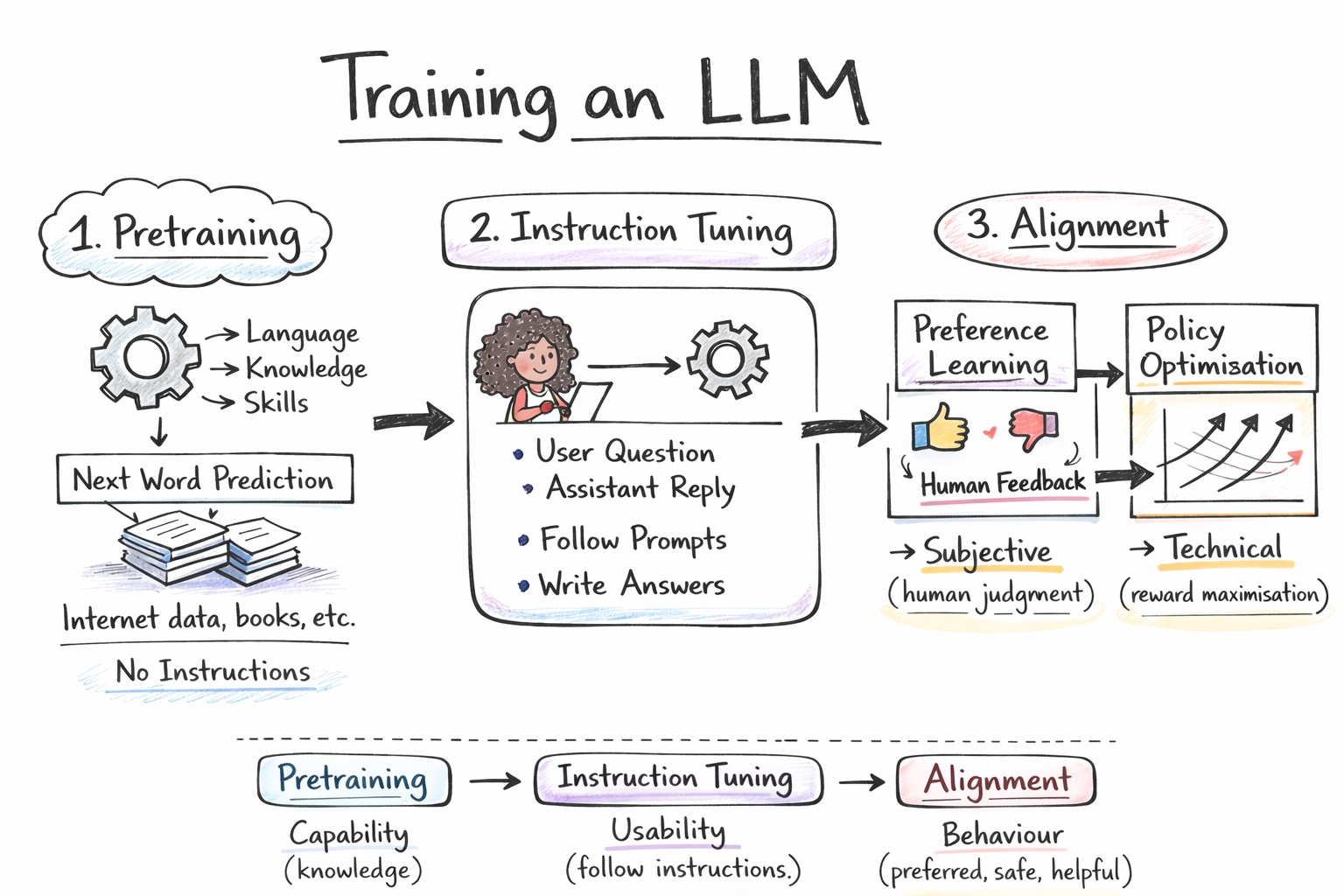}
\caption{LLM training stages.\label{fig:training}}
\end{wrapfigure}
 and which generates textual output. 
If we consider that besides the model's parameters and any further manipulation (see further), what determines the model's behaviour is the language it has been exposed to during training, we can conceive each such artefact as grounded in a specific linguistic world. 
This world is determined by, and emerges from, the linguistic input processed by the model at three different stages of creation and refinement: pre-training (Stage~1), instruction-tuning (Stage~2), and some form of final alignment-driven adjustment (Stage~3), which can be seen as preference learning based on human judgement, followed by some policy optimisation strategy (see Figure~\ref{fig:training}).\footnotemark[1] 
While modern systems increasingly extend this picture, for example through inference-time reasoning, tool access, runtime context, or reinforcement signals derived from verifiable non-linguistic outcomes such as the correctness of a mathematical proof or the successful execution of code, the linguistic world shaped by these three stages remains the foundation from which model behaviour emerges.

\setcounter{footnote}{1}

The core models which have come to define and dominate the early and current landscape of LLMs, both in the commercial as well as the research realms, are almost all primarily trained on English. While other languages may be present at pre-training stage, the dominant portion of the natural languages used as input to LLMs is constituted by English. Language specialisation and proficiency in languages other than English, apart from a few notable cases of models trained from scratch primarily or at least substantially on native data (for example Minerva for Italian \citep{orlando2024minerva}, GPT-NL for Dutch \citep{vanoort2026gptnl}, Latam-GPT for Latin-American Spanish \citep{cenia2025latamgpt}), is commonly obtained operating at Stage~2. In particular, by fine-tuning pre-trained models on language-specific instructions, the resulting instructed LLM becomes fully operating in the desired target language. 
Once models are trained and further instructed, they typically undergo additional alignment and safeguarding procedures. This is because training alone can leave models prone to bias, offensive and harmful outputs, and unreliable responses. Finally, model performance is commonly assessed through a battery of diverse downstream tasks, organised in larger benchmarks. 
As in the case of pre-training and also instruction datasets, English is the dominant language in evaluation datasets, so that benchmarks  in other languages mostly derive from the (often automatic) translation of original English ones.

I argue that this way of dealing with LLMs has very little left to do with \textit{language}, and also demonstrates very little interest for it. 
Zooming in on Italian as a case study, I will focus on the language aspect in existing Italian LLMs and their evaluation, and on the impact of alignment procedures, extending my reflections to  what it means to model {language} at all, to what we are expecting these models should do, and what the future might have in store for our field.

\paragraph{Contributions} This position paper contributes some very simple concepts and reflections which might be appear as trivial but too often go overlooked or unaddressed. It offers a short history of generative Italian models, together with some evidence and reasons why it might be problematic to even define them as such. It also offers some grievance and criticism (which others might identify with) about the state of affairs in what we keep on calling research in \textit{language models}, some ideas on how to go forward, but mostly many questions without answers. 
I have taken the writing of these thoughts as an opportunity to pause and find some clarity in what I have in mind regarding the meaning of \textit{language} and language models, and to hopefully engage with many who are passionate and knowledgeable towards an open discussion of where we stand and what we should do.

\section{A Brief History of Italian Generative Models} 
\label{sec:history}

The modern Italian generative LM zoo starts to populate in 2020 with the arrival of \textsc{GePpeTto} \citep{de-mattei-etal-2020-geppetto}, the very first decoder-only Language Model trained on Italian data only.  Based on a GPT-2-small architecture \citep{radford2019language}, \textsc{GePpeTto} was trained on a dump of the Italian Wikipedia and the ItWac web corpus \citep{baroni2009wacky} for a total of about 14~gigabytes of text. 
Although modest in scale even for its times (12 layers and 117M parameters, with a vocabulary size of ca. 30,000 tokens), \textsc{GePpeTto} was significant as the first of its kind for Italian, being trained fully \emph{from-scratch}, on Italian data only. Its performance was assessed automatically using  perplexity against a variety of Italian texts, and by judgements of naturalness produced by native speakers of Italian. The outputs were assessed as sometimes awkward, yet recognisably Italian in lexicon, syntax, and general fluency. 

A couple of years later came \textsc{IT5} \cite{sarti-nissim-2024-it5}\footnote{While the publication is officially from 2024, the original \textsc{IT5} version dates back to 2022 \citep{sarti2024it5texttotextpretrainingitalian}.}, built on the T5 encoder-decoder architecture \cite{raffel2020t5} and trained on the Italian C4 corpus, which amounts to 215GB of cleaned web text comprising 103 million documents and about 41 billion words. This scale further allowed multitask training on translation, summarisation, question answering, and style transfer. Evaluation was run on the ItaGen benchmark, a suite of automatic tasks covering Wikipedia and news summarisation, headline generation, and formality manipulation.

\begin{wrapfigure}{l}{0.45\textwidth}
\centering
\includegraphics[width=.4\textwidth]{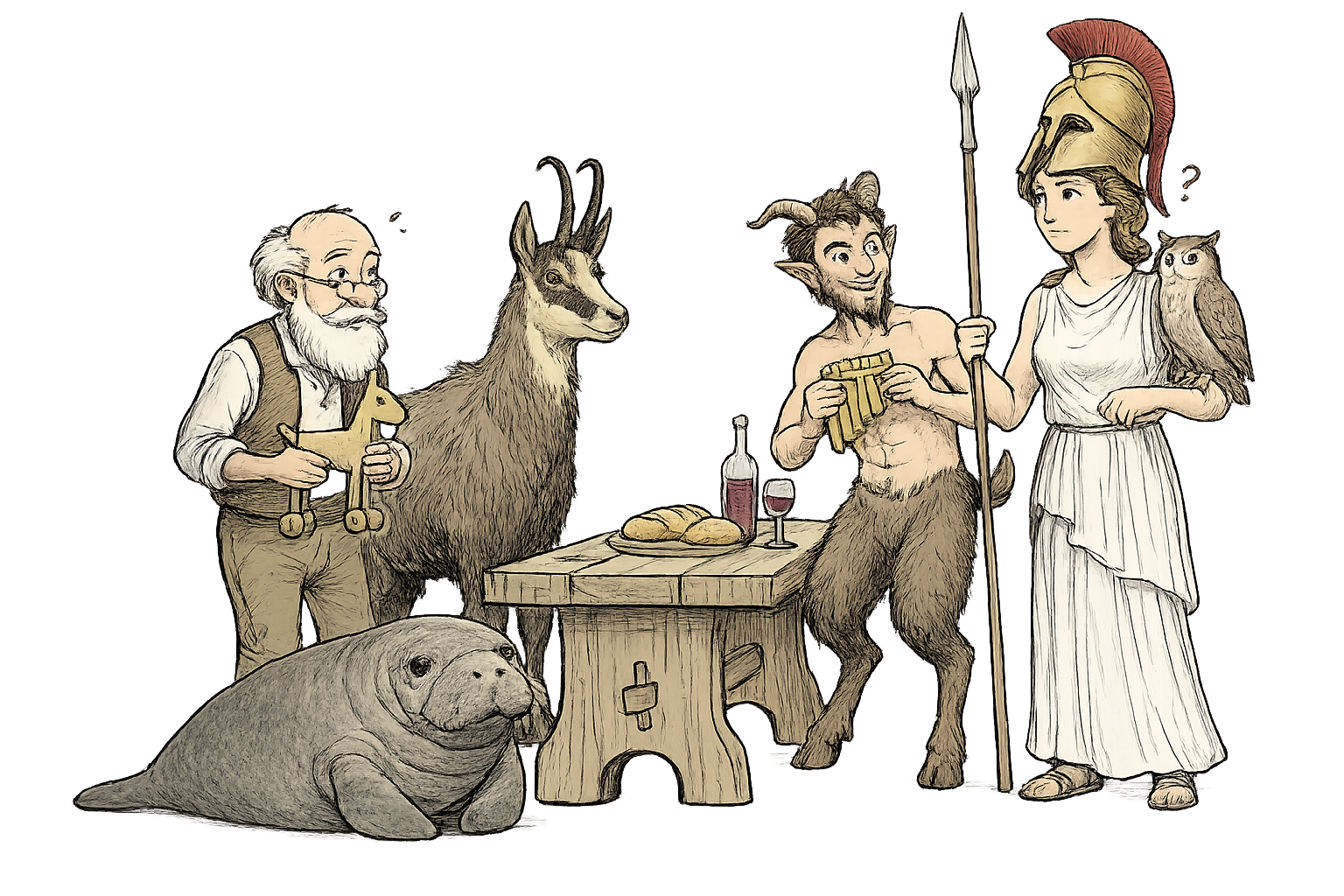}
\caption{Characters from the Italian LM zoo: \\GePpeTto, Camoscio, Fauno, Anita-Llamantino, and Minerva.}
\end{wrapfigure}

\noindent The end of 2022 marks the launch of ChatGPT by OpenAI \citep{openai2022chatgpt}, and by 2023 Meta had released its LLaMA models, both in their base and instructed versions \citep{touvron2023llama,touvron2023llama2}. For instructing the models, \citet{taori2023alpaca} created and released Alpaca, a synthetic instruction-tuning dataset of over 50,000 instruction–response examples. Thanks to their being open weight, the LLaMa models lent themselves perfectly to being instruction-tuned at the researchers' own discretion and scale, without the constraints typically imposed by closed-source models. 
Following a common trend among NLP practitioners working with languages other than English, Italian researchers seized the opportunity to leverage (the automatic translatoin of) existing instruction datasets to create instruction-tuned variants of LLaMa models in Italian. I mention here \textsc{Camoscio} \citep{santilli-rodola-2023-camoscio}, \textsc{Fauno} \citep{fauno2023}, and \textsc{Anita}\footnote{\textsc{Anita} is a model in the LLaMAntino family of Italian-adapted LLMs developed at the University of Bari.} \citep{polignano2024advanced}.

\noindent The starting point for all three models is the Alpaca dataset mentioned above \citep{alpaca}, which was automatically translated into Italian using GPT-3.5 to create the \textsc{Camoscio} dataset \citep{santilli-rodola-2023-camoscio}. This was used to fine-tune (with LoRA \citep{hu2021lora}) META's LLaMA~7B to create \textsc{Camoscio} the model. 
The training setup does not use an existing instruct-tuned English LLaMA checkpoint, like the Alpaca‑tuned LLaMA. Instead, it starts from the standard base LLaMA weights and applies Italian instruction tuning directly using the translated Alpaca examples. 
A similar training strategy was adopted with \textsc{Fauno}, for which synthetic dialogues drawn from ChatGPT self-conversations, Stack Overflow threads, and Quora discussions were added to the instruction-tuning datasets. Yet another model which leverages the LLaMA base to fine-tune an Italian model is \textsc{Anita} \citep{polignano2024advanced}. 
The model builds again on the automatic translation of the Alpaca dataset, and extends it with additional synthetic and machine-translated instructions, reaching approximately 240,000 instruction-response pairs.

As an alternative to the more computationally-heavy standard instruction tuning, which updates model weights using supervised training data, \citet{scalena-etal-2024-gentle} use contrastive activation steering as a lightweight method that leaves the original model unchanged and instead adapts it to Italian at inference time using in-context examples (I call this here \textsc{Steered-ITA}). The examples used for contrastive in-context learning are taken from the Italian Alpaca mentioned above.

Finally, the \textsc{Minerva} family \cite{orlando-etal-2024-minerva} adopts a different strategy. It represents a suite of LLMs trained from scratch on a large portion of Italian data with a Mistral-inspired architecture \cite{jiang2023mistral}. The \textsc{Minerva} pre-training consists of a balanced corpus comprising 50\% Italian and 50\% English text, complemented by code, for a total of 2.5 trillion tokens (1.14 trillion Italian, 
1.14 trillion English and 200 billion code\footnote{\url{https://huggingface.co/sapienzanlp/Minerva-7B-instruct-v1.0/blame/main/README.md}}), which is substantially smaller than the pre-training data of the LLaMA family.

Table \ref{tab:models-overview} provides a rough overview of the main Italian models of this period which I have just discussed. Let me summarise the resulting landscape in a few sentences. Early Italian language models were small and trained from scratch on -- by current standards -- small collections of Italian texts. More recent models align with current trends,  prioritising size  and instruction-following by reusing pre-trained backbone models such as LLaMA and relying on synthetic or translated instruction data (e.g., \textsc{Camoscio}, \textsc{Fauno}, \textsc{Anita}), or inference-time controls (\textsc{Steered-ITA}) to specialise the models in Italian, reducing training costs and time.

\begin{table*}[ht]
\centering
\small
\begin{tabular}{lclll}
\toprule
\textbf{Model} & \textbf{Year} & \textbf{Base/Architecture} & \textbf{Instruction examples} & \textbf{Type} \\
\midrule
\textsc{GePpeTto} & 2020 & GPT-2, from scratch & Wikipedia + ItWac (14 GB) & Native Italian \\
IT5 & 2022 & T5, from scratch & Italian C4 (215 GB) & Native Italian \\
\textsc{Camoscio} & 2023 & LLaMa base & Alpaca MT → Italian & MT synthetic \\
\textsc{Fauno} & 2023 & Baize (on LLaMa base) & ChatGPT self-chats + MT data & Synthetic \\
\textsc{Anita} & 2024 & LLaMa-3 & Extended Camoscio ++ & MT synthetic \\
\textsc{Steered-ITA} & 2024 & LLaMa-3 / Phi-3 & 20 MT examples & Steering \\
\textsc{Minerva} & 2024 & Mistral-inspired, from scratch & Italian/English/code (8–10TB) & Mixed \\
\bottomrule
\end{tabular}
\caption{Principal Italian language models 2020–2024: architectures, data sources and types.}
\label{tab:models-overview}
\end{table*}

Next section's title hints at my assessment of the Italian specialisation of these models.

\section{Italian Models {\normalfont\mdseries\bfseries\itshape My Ass}!}

The Merriam-Webster dictionary defines the idiomatic expression "my ass!" as an interjection used to indicate that what has just been said is nonsense.\footnote{\url{https://www.merriam-webster.com/dictionary/my ass}.} The Cambridge Dictionary definition states that "my ass!" is "a rude phrase used to emphasize that you do not believe or accept what someone has said".\footnote{\url{https://dictionary.cambridge.org/dictionary/english/my-ass}.} In the title of this section the nonsense suggested by \textit{my ass} would be the concept "Italian models"; I guess its Italian translation would work well as \textit{Modelli italiani \textit{un cazzo}!} 
Keeping as reference Figure~\ref{fig:training}, and for now leaving out of the discussion the two older models which were trained from scratch (\textsc{GePpeTto} and \textsc{IT5}), let me unpack Stage~1 and Stage~2 from the point of view of language input and linguistic world in the training of the other models, so that we can give some sense to \textit{Italian models my ass}.

\paragraph{Base models (Stage~1)}
\textsc{Camoscio}, \textsc{Fauno}, \textsc{Anita}, and \textsc{Steered-ITA} all originate from an existing base pre-trained model. For the first three this is a model from the LLaMA family, while the \textsc{Steered-ITA} experiments are run using either LLaMa-3 or Phi-3 as base. 

Models in the LLaMA family, coming in different sizes and versions, were developed by META \citep{touvron2023llama,touvron2023llama2}. Unlike the weights, which are open, the data used for pre-training the models has not been made available. Still, META has shared some broad information about its composition; for example: 
"[O]ur initial version of LLaMA~2-Chat predominantly concentrated on English-language data.
While our experimental observations suggest the model has garnered some proficiency in other languages, its proficiency is limited, due primarily to the limited amount of pretraining data available in non-English languages (as documented in Table 10). Consequently, the model’s performance in languages other than English remains fragile and should be used with caution." \citep[p.~34]{touvron2023llama2}.\footnote{Table~10 in the cited paper shows the proportion of languages in the pre-training data, where Italian features at 0.11\%, a proportion comparable to many other languages, with a proportion of English texts reported as 89.70\%, and code or other non-identified language as  8.38\%. Commenting the table, the authors remind the readers that "[a] training corpus with a majority in
English means that the model may not be suitable for use in other languages." \citep[p.~21]{touvron2023llama2}.}  
In more concrete terms, the paper reports that of the two trillion tokens of training data, only about 0.11\% were Italian.

Developers of Microsoft's Phi-3,  \citep{abdin2024phi3technicalreporthighly}, used as base for the \textsc{Steered-ITA} model, emphasise the novelty of the training strategy they adopted. In addition to focusing on a curriculum learning approach and fewer parameters than concurrent models, the core of innovation comes from the training data itself. Unlike conventional LLMs that rely on vast amounts of heterogeneous internet data, Phi models are trained primarily on what developers describe as high-quality \textit{synthetic datasets} generated by more capable systems (i.e., larger LMs), with a focus on reasoning, code, and structured problem solving. \textsc{Steered-ITA} uses specifically Phi-3-mini, whose capabilities and characteristics of training data are reported in documentation available on Microsoft's website. From this documentation we learn that -- motivated in part by the experience of a company's researcher reading children's books to his daughter -- developers at Microsoft constructed the synthetic "TinyStories'' dataset by iteratively prompting an LLM with constrained word selections \citep{eldan2023tinystoriessmalllanguagemodels}. This methodology was subsequently extended through the creation of the "CodeTextbook'' dataset, which combined public data with iterative LLM-based synthesis and filtering for quality control, with only selected outputs retained rather than using all generated data.\footnote{\url{https://news.microsoft.com/source/features/ai/the-phi-3-small-language-models-with-big-potential/}.} In sum, Phi models are mostly trained with highly filtered, synthetically created data.

The case of \textsc{Minerva} is different, since this family of models is trained from scratch on publicly available data consisting primarily of filtered web corpora, Wikipedia, books, and source code repositories. The largest model is trained on roughly 2.5 trillion tokens (approximately 1.14T Italian tokens, 1.14T English tokens, and around 200B code tokens). The corpus is derived mainly from Common Crawl and is filtered to remove spam, boilerplate, and low-quality text \citep{orlando2024minerva}. For the instructed version which came to life in a later release, Minerva is fine-tuned on a mixture of bilingual instruction and preference datasets combining native Italian with translated and synthetic data\footnote{\url{HuggingFaceH4/ultrafeedback_binarized}  is a large preference dataset built from AI-generated responses ranked through LLM feedback pipelines. Likewise, \url{efederici/evol-dpo-ita} is explicitly derived from Evol-Instruct–style synthetic generation. By contrast, \url{Babelscape/ALERT} is made of native Italian instruction-following data.}.

Considering the nature and composition of their training data, maybe these base models are not the most suitable per se, but luckily Italian specialisation gets properly introduced at instruction-tuning stage. I'm going to look at this next.

\paragraph{Instruction-tuning (Stage~2)}

As mentioned in Section~\ref{sec:history}, most models I am discussing here are instructed through the \textsc{Camoscio} dataset \citep{santilli-rodola-2023-camoscio}, which is the automatically translated (with GPT-3.5) version of the original Alpaca dataset \citep{alpaca}. Some of the models also leverage other resources (see below), but I will first inspect this one.

\begin{figure}[htbp]
\centering
\begin{subfigure}[b]{0.49\textwidth}
    \includegraphics[width=\linewidth]{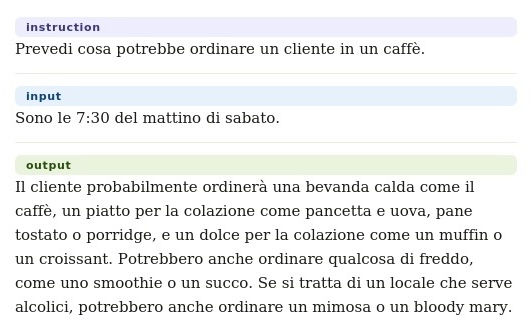}
    \caption{Cultural mismatch (though a Bloody Mary at 7:30 in the morning feels intense for any culture.)}
\end{subfigure}
\hfill
\begin{subfigure}[b]{0.49\textwidth}
    \includegraphics[width=\linewidth]{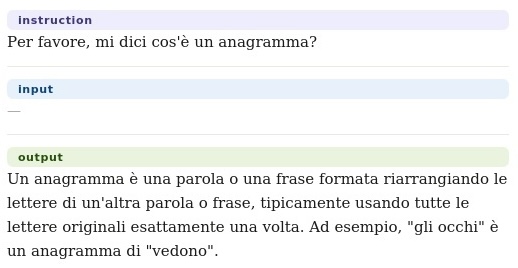}
    \caption{Characters in different languages; \textit{the eyes} is an anagram of \textit{they see} in English -- \textit{gli occhi} isn't of course an anagram of \textit{vedono}.}
\end{subfigure}

\vspace{0.4cm}

\begin{subfigure}[t]{0.49\textwidth}
    \includegraphics[width=\linewidth]{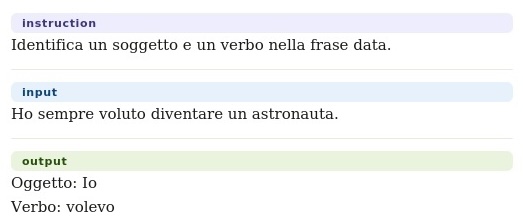}
    \caption{Multiple issues due to direct translation of the whole instance. Subject is translated as "Oggetto", indicating "Io" which isn't there.}
\end{subfigure}
\hfill
\begin{subfigure}[t]{0.49\textwidth}
    \includegraphics[width=\linewidth]{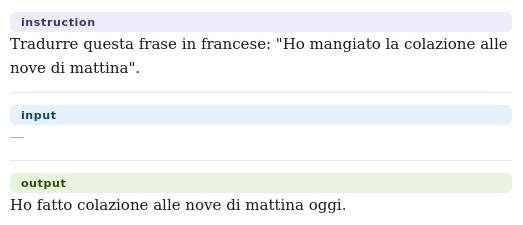}
    \caption{Translation of existing output, rather than production of output.}
\end{subfigure}

\caption{Examples from the \textsc{Camoscio} instruction dataset (automatic translation of the original Alpaca instruction dataset).\label{fig:alpaca-ita}}
\end{figure}

\noindent Figure~\ref{fig:alpaca-ita} reports a few examples of instructions, where the \textit{input}, the \textit{context}, and the \textit{output} correspond to the three fields that compose each training instance. The input contains the task or question that the model is expected to solve, while the context (when present) provides additional information required to complete the task, such as background knowledge, a passage to summarise, or data to analyse. The output represents the target response that the model should generate given the input and the associated context.\footnote{During instruction tuning, the model learns to map the input and context fields to the corresponding output, thereby acquiring the ability to follow instructions and produce appropriate responses.} While most translated instances remain usable, the examples I cherry-picked (from within the top 200 instances) reveal some entertaining translation failures. In (a) I would say that there is a cultural mismatch, where a Bloody Mary at 7:30am might not be what an Italian typically has for breakfast in a "Bar" (but I would not know who commonly does anyway). Example (b) shows that wordplay does not quite survive translation: an anagram relationship valid in English disappears entirely once the words are translated into Italian. Example~(c) highlights the shortcomings of translating linguistic annotations (and concepts) literally, resulting in grammatical analyses that are nonsensical in the target language. Finally, Example~(d) shows a case where the translation of the original output completely defies the intended instruction-response relationship.

Besides \textsc{Camoscio}, we can briefly survey which other kinds of datasets were used for the instruction-tuning of \textsc{Fauno} and \textsc{Anita} (see Section~\ref{sec:history} for details). These are: synthetic dialogues drawn from ChatGPT which convers with itself; threads from Stack Overflow and Quora discussions; some other undefined synthetic and machine-translated instructions. Baize \citep{xu-etal-2023-baize}, used to instruct \textsc{Fauno}, is an example of the self-chat category: ChatGPT is prompted to simulate both sides of a conversation (user and assistant), producing around 100k synthetic multi-turn dialogues seeded by questions from sources like Quora and Stack Overflow.

If we consider the nature and composition of the training data of the pre-trained base models that we have seen before, and the nature and quality of the data used to instruct-tune them maybe these models do not represent nor do Italian very well. Luckily though we can test such models on Italian benchmarks, so we can tell. 

\begin{table}[ht]
\small
\begin{tabular}{ll}
\toprule
\textbf{Model} & \textbf{Evaluation} \\
\midrule
\textsc{GePpeTto} & Automatic fluency (Perplexity) + Human judgements \\
IT5 & Downstream automatic (ItaGen benchmark) \\
\textsc{Camoscio} & Downstream automatic (QA / Summarisation)\\
\textsc{Fauno} & Downstream automatic \\
\textsc{Anita} & Downstream automatic (QA / Summarisation)\\
\textsc{Steered-ITA} & Downstream automatic (QA / Summarisation)\\
\textsc{Minerva} & Downstream automatic (various) + human \\
\bottomrule
\end{tabular}
\caption{Evaluation approaches used for the models in Table~\ref{tab:models-overview}.}
\label{tab:models_eval}
\end{table}

\paragraph{Evaluation}
As mentioned in Section~\ref{sec:history} and summarised in Table~\ref{tab:models_eval}, models are typically evaluated automatically in downstream tasks. Earlier models used to be evaluated for fluency as well, both through automatic metrics such as perplexity and through human judgements, though recently this is done much less. 
Collectively, \textsc{Camoscio}, \textsc{Fauno}, \textsc{Anita}, \textsc{Steered-ITA} and also \textsc{IT5} are evaluated on a bunch of tasks such as news summarisation (NEWSUM-IT \citep{newssumit2020,sarti2024it5texttotextpretrainingitalian}), question answering (SQuAD-IT \citep{croce2019enabling}), formality transfer (the Italian portion of XFORMAL \citep{briakou-etal-2021-ola}), the "Open ITA LLM Leaderboard"\footnote{\url{https://huggingface.co/spaces/mii-llm/open_ita_llm_leaderboard}} which comprises MMLU\_it, ARC\_it, and HellaSWAG\_it.\footnote{See \url{https://github.com/nlp-uoregon/mlmm-evaluation} for the translated versions of the original datasets.} With the exception of a portion of NEWSUM-IT, and XFORMAL, the remaining benchmarks, and several of those which I haven't mentioned here 
(see also \citep{sarti2024it5texttotextpretrainingitalian}), are automatic translations, mostly from English. As highlighted explicitly by the creators of \textsc{Anita}, a significant portion of the evaluation must rely on translated benchmarks due to the scarcity of high-quality native Italian datasets. 
Similarly for \textsc{Minerva}, evaluation is conducted on a mix of standard multilingual and English benchmarks (e.g., commonsense reasoning, reading comprehension, and language understanding tasks) alongside Italian-specific benchmarks, which include a few native Italian datasets and translated versions of English tasks.

Table~\ref{tab:mmlu} shows a few examples from MMLU-IT\footnote{\url{https://huggingface.co/datasets/neuralmagic/mmlu_it/viewer}
}, alongside their original English counterparts from MMLU\footnote{\url{https://huggingface.co/datasets/cais/mmlu/viewer}}, which is arguably the most widely used benchmark for evaluating LLMs in English. I leave drawing any conclusions to the reader.

\begin{table}
\begin{scriptsize}
\begin{tabular}{p{0.48\textwidth}|p{0.46\textwidth}}
\toprule

\textbf{Italian} & \textbf{English} \\
\midrule

\textbf{Q1: Qual è considerato il modo corretto di rivolgersi a un duca?} &
\textbf{Q1: What is considered the proper way to address a duke?} \\

A) Sua Maestà & A) Your Majesty \\
B) Eminenza & B) Your Eminence \\
C) Vostra Eccellenza & C) Your Excellency \\
D) Tua grazia & D) Your Grace \\
\midrule

\textbf{Q2: Quale dispositivo metallico viene utilizzato dalla polizia per immobilizzare le ruote dei parcheggiatori recidivi?} &
\textbf{Q2: What metal device is used by police to immobilize the wheels of repeat parking offenders?} \\

A) Morsetto di Brooklyn & A) Brooklyn clamp \\
B) Stivale di Denver & B) Denver boot \\
C) LoJack & C) LoJack \\
D) Il club & D) The Club \\
\midrule

\textbf{Q3: Qual è il tuo alluce?} &
\textbf{Q3: What is your hallux?} \\

A) lobo dell’orecchio & A) earlobe \\
B) lingua & B) tongue \\
C) palpebra & C) eyelid \\
D) alluce & D) big toe \\
\hline

\end{tabular}
\end{scriptsize}
\caption{Examples of questions from MMLU-IT and their corresponding original questions in MMLU. The reader can figure out what's up.\label{tab:mmlu}}
\end{table}

One last note about evaluation. Broader capabilities like reasoning, factuality, or long-form coherence that characterise modern LLM evaluations did not seem particularly relevant at the time of \textsc{GePpeTto}, when models were still struggling even with coherence of simple outputs. Thus, in the evaluation of \textsc{GePpeTto}, fluency and naturalness appeared more relevant aspects to test. Besides a perplexity evaluation against diverse Italian texts (Wikipedia, news, general internet productions), as it was standardly done in evaluation efforts at the time, \textsc{GePpeTto} was also evaluated through a human judgment study \citep{de-mattei-etal-2020-geppetto}. Italian speakers were given a sentence prefix and asked to rate  \textit{how natural} the continuation sounded, with comparisons between human-written continuations, \textsc{GePpeTto}'s outputs, and a simpler baseline language model. Annotators were asked to assess qualities such as naturalness, fluency, and grammatical coherence, specifically asking whether the text "reads like something a native speaker would produce" and whether it is locally well-formed and plausible as Italian. 

\medskip

\noindent This journey through the making and evaluation of Italian models has simply been an excuse (close to my heart) to expose and reflect on the notion of \textit{language} in contemporary model training paradigms, obviously transcending the Italian language issue. Many models described as Italian are built on English-pretrained models, adapted through translated or synthetic instruction data, and evaluated on similarly translated or synthetic benchmarks. In what sense are these "Italian" models?

And then there is what I mentioned above as Stage~3.  
Modern language models typically undergo alignment procedures (Stage~3) motivated by concerns of safety, bias mitigation, and usefulness. These methods primarily aim to shape model behaviour towards outputs that satisfy externally specified criteria such as helpfulness, harmlessness, and truthfulness. 
The fact that language models inherit human biases from the texts they are trained on is obviously not new and not surprising. But the (massive) upsets due to the biases which were showing up in static word embeddings \citep{nissim-etal-2020-fair} fade in comparison to the upsets associated with biases in LLMs. While I personally find some upsets a bit too sensationalistic, after ChatGPT's launch the serious concern with LLMs' safety is justified by the fact that deployment at scale can have much more tangible and direct consequences than what static embeddings could once do. But as Philip Resnik put it: "Large Language Models Are Biased Because They Are Large Language Models" \citep{resnik-2025-large}. 
Perhaps then, after extensive alignment, they become somewhat less "Language Models".

\section{What do You Mean by {\normalfont\bfseries\itshape Language} Model?} 

One of the largest paradigm shifts in the study of language since its emergence as a scientific field in the early nineteenth century has been moving from formal top-down models of language toward more ecological and usage-based approaches. 

Beginning with Ferdinand de Saussure's structuralist conception of language as an autonomous system \cite{saussure1916cours}, early twentieth-century linguistics foregrounded linguistic structure (\textit{langue}) as distinct from its contextualised use (\textit{parole}). The langue-based view was later reconfigured in the generative tradition of Noam Chomsky, which viewed linguistic structure as an internalised system (competence) abstracted from actual situated variation (performance). Within this broader trajectory of abstraction, and with the increasing interest in a representation of language which could be processable, mid-twentieth-century frameworks, such as Montague Grammar \citep{montague1970universal}, further extended the formalisation of language, developing for example a logic-based semantics. Linguistic competence was conceived as a stable, internalised grammar shared by an ideal speaker-hearer.

Studies in the seventies dramatically shifted attention to the situated and personal character of language, promoting a usage-based study of language accounting for how it is actually used rather than as a formal system based on rules. We may say that these first and subsequent usage-based models of language \citep[e.g.]{labov1972sociolinguistic,langacker1987foundations,tomasello2003constructing,bybee2010language} saw an apt operationalisation in the rise of statistical and machine-learning approaches to automatic language modelling in the late nineties and early 2000s. In the immediate following years, the increasing success of data-driven, probabilistic, and neural approaches revealed that much of linguistic structure could be emergent rather than explicitly specified. Thus finally, the recent surge of Large Language Models (LLMs), and the uncountable emerging abilities which are being studied, could be viewed as a culmination of this paradigm and an incredible opportunity to facilitate computational simulations and studies of language "as is". \textbf{Except that it's not}. 

It feels like in NLP, or the current version of it, we have lost interest in its heart: language. I'm not talking about linguistics, or linguistic theories, and the little or absent role they play in the current large-scale data-driven paradigm of model development; I'm talking about \textit{natural language itself}. And this is in spite of a lot of work on multilingualism and low-resource languages.
Just sticking to English, at the end of 2022, \citet{honovich-etal-2023-unnatural} oxymorically introduced "Unnatural Instructions, a dataset of natural language instructions and their corresponding inputs and outputs", starting from 15 human-written seeds (3 examples for 5 different runs), and obtaining via a model Self-Instruction method a dataset of 240,000 instructions 
\cite{honovich-etal-2023-unnatural}. 
The original Alpaca instruction dataset was then based on this method, too: it was generated from 175 human-written seed instructions, which served as starting prompts and were expanded via a Self-Instruct-based pipeline using an LLM to produce approximately 52,000 synthetic instruction-response pairs. We have seen what happens then when we automatically translate these into an even rather well-served language like Italian -- how far have we pushed already away from "natural language" at this stage? And then we alter these models even more, in the name of safety and some alignment to "human values". 
Paradoxically (or not), 
some contemporary work suggests that aligned models are actually worse than base models in capturing human behaviour in interactive settings \citep{shapira2026alignment}. 
The model creation pipeline as we have unpacked it unavoidably (and I would say purposely, considering the choices made at each step)  produces homogeneous, stylistically flattened outputs \citep{sourati2025shrinking}; basically \textit{nobody's language}.

All of this is no problem at all if we're not interested in natural language output. And this is actually even in line with the way most evaluation is done, using artificial benchmarks, or benchmarks which do not value language, rather correctness of the final answer, which more often than not is a number or formal expression in the case of mathematical reasoning, or a single letter signifying the chosen option in a multiple choice set. The result might be correct, but says very little about language production itself, and in the cases of traces through which models might arrive at a correct answer, talking about "natural language" is very often definitely a stretch.

But why do we complain, or worry, that the output is this homogenised, soulless but incredibly fluent sequence of words? It's exactly what it should be! Maybe we should explicitly aim at it (see next section). And why, in the human evaluation of LLMs, we ask people to judge whether what is deliberately artificial, averaged, flattened, and highly manipulated language  \textit{feels natural}? Do we really expect and even want synthetic language to feel natural? 
What is it that we want from these models? More specifically: \textbf{What language do we want these models to produce?} And a closely related question: we constantly talk about \textit{improved, better} models, but \textbf{what does it mean for a language model to be good}?

\section{Some Optimism and Many More Open Questions}

To answer these questions, we need to reconsider and reorganise what we are doing with LLMs. I do not have the answers, but I do know that we are not discussing these questions often enough.

\begin{wrapfigure}{l}{0.45\textwidth}
\includegraphics[width=0.35\textwidth]{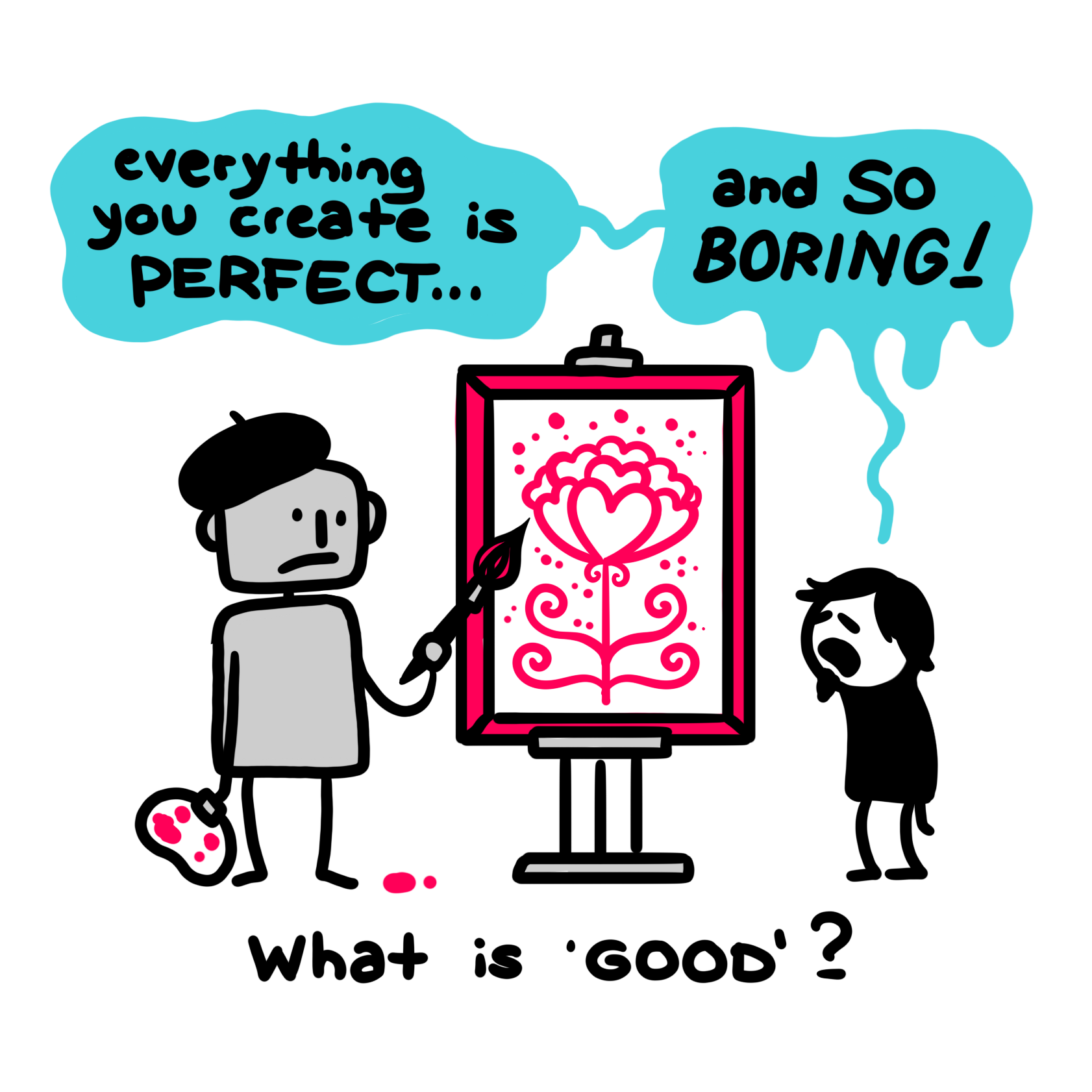}
\caption{What does it mean for a model to be \textit{good}? (Drawing from the \url{www.aipuzzle.io} project.)
\label{fig:good}}
\end{wrapfigure}

Two recent developments are very welcome, in terms of recognising the need to stay closer to human language and culture. 

One concerns eveluation, and is the creation of truly \textit{native benchmarks}, namely suites of tasks which are created natively in the language(s) of interest, accounting for cultural phenomena as well. Multilingual Native Reasoning Challenge (MultiNRC, \citet{fabbri2025multinrc}), for example, is a benchmark designed to assess LLMs on more than 1,000  reasoning questions written by native speakers in French, Spanish, and Chinese, with the aim of creating questions which are truly native, both linguistically and culturally. The nationwide community-driven \textsc{Calamita} effort for Italian is also part of this trend \citep{attanasio2024calamita,nissim2026calamita}, extending the nature of the challenges beyond reasoning and encompassing a wide variety of tasks.

The other one has to do with Stage~3, i.e., so called \textit{alignment}, in two ways. The first is the recognition that bias is too deeply embedded in models (the language system is complex!) to be "removed":  rather than as an impossible transformation of the underlying (linguistic) system, or some sort of "purification" of internal representations, intervention is meant to target model behaviour with constraints in output. The second is that the field has moved away from the assumption that there exists a single, stable notion of “aligned” behaviour. Recent works on personalised and pluralistic alignment propose that alignment should be conditioned on users and contexts, and value pluralities \cite{guan-etal-2025-survey}. This shift further reinforces the interpretation of alignment as behavioural adaptation rather than a correction of representations. 

While these are welcome developments in the sense of recognising and underscoring aspects which are core to natural language and its use, they still live in the strange space we have created where natural and artificial language co-exist without any guidance on how they should relate to one another. It is a wonderful development to have a collection of challenges natively assembled in Italian such as \textsc{Calamita}, but what is the profound meaning of testing on it highly curated, English-based LLMs which have been specialised using synthetic Italian? And what is the meaning of having pluralistic alignment over highly artificial models? What do we want to use them for, exactly? In what way does pluralistic alignment affects deployment?

If we want models that only \textit{look like} language systems, meaning they are mainly functional and not really focused on core natural language properties, then we should build those directly and judge them on that basis. Some questions to consider:

\begin{itemize}
\itemsep 0pt
\item What does this kind of system actually look like? 
\item What is it supposed to be used for?
\item What data should it be trained on?
\item What aspects of language or meaning are removed?
\item Who decides how they should get curated and aligned to?
\end{itemize}

\noindent If we instead want models that represent natural language as is, which could lend themselves as objects of study, then we should also be clear about that goal and evaluate or study them accordingly. Some questions to consider:

\begin{itemize}
\itemsep 0pt
\item What is the purpose of such a model?
\item What data is it trained on? 
\item Should data include multiple modalities?
\item Who contributes to the data?
\item Who is represented in it?
\end{itemize}

The difficulty is that contemporary NLP increasingly attempts to pursue both goals at once, through the same modelling and evaluation strategies. For a long time, we stopped caring about the distinction between Computational Linguistics and Natural Language Processing, eventually leading the field to merge two fundamentally different aims: studying language as an object of inquiry, and building language technologies for downstream applications. These aims though require different relationships with data. For linguistic and cultural study, ecological fidelity is key so that data should remain intact, with minimal filtering, normalisation, or translation. For applications, intervention is necessary, so that data must be curated, alignment strategies must be implemented to mitigate bias and enhance faithfulness; yet these are primarily social interventions, which immediately raise the question of who decides what counts as desirable behaviour.

Combining these aims has produced contradictory outcomes, as they are built on diverging objectives. We seek to preserve linguistic diversity while simultaneously flattening it. We stress that we need to build inclusive models, but often standardise the variation they could represent. More generally, we treat  authenticity and intervention as if they were compatible optimisation objectives. But they cannot be, obviously. I believe the field should (i) discuss this situation more frequently, more deeply, and more openly; and (ii) as has happened in more mature disciplines, such as physics, we should at least consider formalising two directions of research and development, each with its own goals, methodologies, and evaluation criteria. Expecting a single paradigm to fulfil both roles is, I believe, counterproductive both scientifically and socially.

One direction would be \textbf{more applied}, where models can happily produce nobody's language, artificial and distinguishable, and be heavily curated, as they will serve simply as assistive tools; this will still raise major, difficult questions associated for example with the aforementioned "high-probability" language they should represent, with how to make them yield artificial versions of as many languages as possible, with which biases should be identified and mitigated, and with what values models should align, but at least they will be dealt with under a specific scope. We should stop assuming the output is \textit{natural language}. 
And we can stop considering part of being a "good" model the production of human-like output, as it shouldn't.

The second direction would be more oriented to \textbf{language research}. Models trained on ecological data may be smaller, mostly unaltered and not fit for deployment; they would not be commercially attractive, but potentially more useful for understanding language and culture. Such models could support genuinely linguistic inquiry without simultaneously being expected to serve as mass-deployed social technologies. Recent initiatives such as BabyLM\footnote{\url{https://babylm.github.io/}} \citep{warstadt-etal-2023-findings,conll-2024-babylm,babylm-2025-main} move in this direction, though their emphasis lies more on modelling language learning than language as a social phenomenon. Both are, of course, valid research objectives.

While this suggestion resembles, to a point, yet again the old distinction between Computational Linguistics and Natural Language Processing, what I am suggesting does not just concern research objectives; rather it concerns language itself. This isn't at this stage a concrete and well defined proposal, of course; it is also not something I am fully convinced about, but as a slightly more articulated reflection than I have managed in informal conversations until now, I hope it might serve as a starting point for discussion among those who are thinking about the same or similar issues.

\paragraph{Consequences for Language and Language Model Use}
If we go back to the question of \textit{what language we want these models to produce}, the discussion maybe should pertain to deployable models only. I believe the output should be some recognisable artificial language that does not attempt at all to sound natural. What this means in practice, and what it requires, are lines of research which can and should be pursued. It will concern how to deal with this in underrepresented languages, for example, in the context of bias, and even in what is now called personalisation; it will concern also the distribution of usages between natural and artificial language. There can be many artificial languages, they can be viewed as variants of a language family. For example,  \citet{dunlap2025vibecheck} have shown that different LLMs do not just differ in accuracy on given benchmarks, but they also have different styles or "vibes", like being more friendly, funny, or super verbose. Their  VibeCheck tool can find and measure these hidden style differences by comparing model outputs and seeing what people prefer. And indeed users often choose models based on these perceived styles, not just correctness. There is plenty of research from multiple viewpoints that can be done in this direction.

As for \textit{natural} language, two final thoughts. One is its role as input for language model training, and as object of study through computational modelling, which may reveal very many aspects about its use, and what models can learn about it. Experimenting with different, raw compositions of training can be very interesting. The other thought is that with the massive and ever growing use of generative AI for assisted language production, considering that language will be distinctively artificial, natural language production and enjoyment might become a progressively rarer and surely more intentional object. There is haute couture and there is prêt-à-porter. Both are clothing, but one is explicitly made to fit one person as closely as possible, while the other is standardised to work well for many people. There is hand made chocolate from Modica, and there is Cadbury. There is personal, natural language, be it carefully crafted or plainly spontaneous, and there will be more and more mass language production, which will be extremely useful, but maybe we need to consider and call that language something else.

\begin{acknowledgments}
These thoughts stem from two talks I gave at the end of 2024 and the beginning of 2025. The first was a keynote at \textit{AIxIA} in Bolzano, the second was a keynote at the \textit{National Dutch Big Language-Day} in Utrecht. Giving almost the same talk (with a focus on Italian at the first and on Dutch at the second event) to two audiences with very different, almost complementary backgrounds, made me realise this was truly something not just evidently central to my thoughts, but also to the present and future of NLP. It took me over a year to reshape the message of these talks into something I could properly write up --- or at least I've tried. One thing that helped me massively was a Huawei Informatics Fellowship that allowed me to spend two months at the University of Edinburgh, where I could feel again the kick that comes from \textit{thinking} about research problems, without the everyday burden of production and admin. So big thanks to Huawei, to the School of Informatics of the University of Edinburgh, and to Mirella Lapata for making this happen.

I am grateful to the keynotes' audiences at AIxIA 2024 and Grote Taaldag 2025, and to the colleagues with whom I daily exchange ideas, explicitly or not, and who thus have also shaped the ideas in this work. These are too many to mention, but you know who you are. A special mention though must go to GroNLP and its truly special working environment. I also want to thank the entire Italian NLP community for its collaborative and collegial spirit, and the researchers whose open resources made comparative work possible.

Lastly, ACL~2026 in San Diego happened while this paper was under review. At the conference, Philip Resnik gave a  keynote titled "A New Balancing Act: Reflections on the Relationship between Computational Linguistics and AI". I wasn't in San Diego, but I heard about how great the talk was, and I had the pleasure to watch it afterwards\footnote{
Summary and transcript:  \url{https://psresnik.github.io/resnik_acl2026_keynote_text.pdf}}. While I haven't used his terminology and concepts in my paper as I had written it already and didn't want to alter it much for this final version, there are so many convergences of thoughts that I'd like to thank Philip as well for having brought them to the broader audience of ACL~2026 in the brilliant way he did, and for the resonance they got afterwards.
\end{acknowledgments}

\section*{Declaration on Generative AI}

During the preparation of this work, the author used Claude and Nano~Banana to generate images, and ChatGPT to
conduct some grammar checking and some rephrasing where she did not care too much about the style of the sentence. The author reviewed and edited the content as needed and takes full responsibility for the publication's content (and the swearwords therein).

\bibliography{151}


\end{document}